# Rapid optimization in high dimensional space by deep kernel learning augmented genetic algorithms

Mani Valleti,[1,2,a] Aditya Raghavan,[1] Sergei V. Kalinin[1,3,b]

[1] Department of Materials Science and Engineering, University of Tennessee, Knoxville, TN 37916 USA
[2]DSSI, NSLS-II, Brookhaven National Laboratory, Upton, NY 11973, USA
[3] Pacific Northwest National Laboratory, Richland, WA

[a] svalleti@bnl.gov
[b] sergei2@utk.edu

## Abstract

Exploration of complex high-dimensional spaces presents significant challenges in fields such as molecular discovery, process optimization, and supply chain management. Genetic Algorithms (GAs), while offering significant power for creating new candidates' spaces, often entail high computational demands due to the need for evaluation of each new proposed solution. On the other hand, Deep Kernel Learning (DKL) efficiently navigates the spaces of preselected candidate structures but lacks generative capabilities. This study introduces an approach that amalgamates the generative power of GAs to create new candidates with the efficiency of DKL-based surrogate models to rapidly ascertain the behavior of new candidate spaces. This DKL-GA framework can be further used to build Bayesian Optimization (BO) workflows. We demonstrate the effectiveness of this approach through the optimization of the FerroSIM model, showcasing its broad applicability to diverse challenges, including molecular discovery and battery charging optimization. The complete set of Pythonic code and datasets to reproduce the results in the manuscript is publicly available in our GitHub repository at https://github.com/saimani5/GA-DKL.

## Introduction

Discovery and optimization in high-dimensional spaces is a fundamental challenge across numerous scientific fields. Typical examples of such problems include molecular discovery, process optimization in materials synthesis and processing, battery charging applications, and manufacturing(1-8). In each of these cases, the goal is to navigate a high-dimensional, differentiable space of potential solutions to identify the most effective strategies or configurations. For example, materials processing(9-13) necessitates discovery of thermal or field profile that gives rise to optimal end properties, for applications ranging from steel and porcelain to battery manufacturing(14, 15). Optimizing battery charging profiles requires defining the optimal current and voltage trajectory that maximizes charging speed(16-21) and minimizes risk of thermal runaway. Manufacturing and chemical engineering often require complex resource allocation and control problems.

Another example is the exploration of high-dimensional non-differentiable spaces, such as molecular spaces(22),(23). In molecular discovery, we seek optimal functionalities within the chemical space of a system. Optimization in molecular spaces is crucial because it enables the discovery of new compounds with desired properties in various fields such as pharmaceuticals, materials science, and energy storage. Efficiently exploring molecular spaces allows researchers to identify novel molecules that can serve as effective drugs, advanced materials with superior properties, or catalysts for chemical reactions. By optimizing molecular spaces, we can uncover hidden relationships and patterns that lead to more efficient and targeted experimentation, reducing costs and time associated with traditional trial-and-error methods. Unsurprisingly, this is now one of the primary research foci for large number of academic and industry research groups.

One of the key methods used for molecular discovery and process optimization are Genetic Algorithms (GAs)(24),(25). Inspired by the principles of natural selection, GAs iteratively evolve a population of candidate solutions towards optimal configurations. The main elements of GAs are mutation and crossover combined with selection(25-27). Mutation introduces random variations to candidate solutions, promoting diversity within the population and aiding in the exploration of the solution space. Crossover, on the other hand, combines segments of two parent solutions to produce offspring, facilitating the exchange of beneficial traits between candidates. The GA typically steps with a selected gene population, applies mutation and cross-over following certain policies to stay in the solution space of the system, and subsequently applies selection algorithms to generate the next generation of genes. While GAs are powerful and robust tools for exploring complex optimization landscapes, they are computationally expensive due to the need to evaluate numerous candidate solutions across many generations. This computational intensity is particularly pronounced in high-dimensional spaces, where the evaluation of all possible trajectories might prove hardly achievable.

The alternative approach for the exploration of the complex spaces involves the use of the classical optimization methods such as Gaussian Process based Bayesian Optimization(28-30). For high-dimensional spaces, these can be performed directly via the choice of proper kernel functions(31), or via the low-dimensional representation of the search spaces. The latter approach leverages techniques like variational autoencoders (VAEs) to reduce the dimensionality of the original high-dimensional search space(32-35). VAEs effectively compress the search space into a latent space that captures the essential features and structures of the data. Once the dimensionality is reduced, GP-BO can be applied to explore this resultant latent space efficiently. Bayesian Optimization, driven by Gaussian Processes, provides a probabilistic model of the objective function, allowing for a more targeted and informed search for optimal solutions. The combination of VAEs and GP-BO offers a powerful strategy for navigating complex optimization problems, as it balances the need for computational efficiency with the ability to thoroughly explore the search space(38, 39). By focusing the optimization efforts on a more manageable latent space, optimal solutions can be identified more effectively and with fewer computational resources. This not only enhances the efficiency of the optimization process but also improves the chances of discovering high-quality solutions that might be overlooked by traditional high-dimensional search methods.

The fundamental limitation of the Variational Autoencoder-Bayesian Optimization (VAE-BO) approach is that the latent space structure is determined solely by the input features and their distribution in the candidate space, but not the target functionality. This means that while VAEs effectively reduce the dimensionality of the search space, the distribution of the target functionality can be very complex insofar it correlates with the features. Consequently, even in simple systems, the target functionality can exhibit highly complex and irregular distributions within the latent space, complicating the BO process. These complexities hinder the efficiency of BO, as it relies on a probabilistic model to guide the search for optimal solutions. Without a well-represented target functionality, the optimization process may struggle to identify high-quality solutions, leading to increased computational effort and suboptimal performance.

As an alternative approach, we have recently demonstrated the use of Deep Kernel Learning (DKL)(40-42) in the context of active learning. DKL combines a neural network for discovering reduced latent representations with a Gaussian Process (GP) that operates over the resultant latent space. In this framework, the GP and neural network are trained jointly after the acquiring a new datapoint, leading to dynamic embeddings of the data rather than static representations as in VAEs which are trained prior to the optimization process on the input dataset. In active learning using Bayesian Optimization (BO), a surrogate model is trained on a subset of input–target pairs to predict target values and quantify associated uncertainties across the remaining input space. The next input to be queried is then selected based on these predictions, making the choice of surrogate model critical to the overall efficiency of the active learning process. Gaussian Processes (GPs) are commonly used as surrogate models in such settings due to their probabilistic nature and ability to quantify uncertainty. However, Deep Kernel Learning (DKL) offers a more flexible alternative by incorporating a neural network in the input layer before the GP layer, enabling richer and more adaptable representations of the input space. This architecture allows DKL to better handle non-stationary and discontinuous functions,(43) accommodate heterogeneous input structures through tailored neural network designs, (41) and suppress input variations in the latent space representations(40) that are irrelevant to the target property.

However, both VAE-BO and DKL approaches have inherent limitations that restrict their use in generative optimization tasks. DKL is not a generative model—it significantly accelerates the exploration of complex search spaces but relies on a fixed set of input examples provided through domain expertise. This poses a key challenge: if the input examples are too broad, the resulting search space becomes intractably large; if too narrow, the optimization may miss high-performing solutions outside the initial dataset. VAEs, on the other hand, are generative and can interpolate between known data points, facilitating broader exploration within the existing data distribution. However, they cannot extrapolate to generate truly novel solutions outside of that distribution. As a result, their capacity to discover new optima is limited to the convex hull of the training data.

In contrast, Genetic Algorithms (GAs) are inherently generative, enabling the creation of new candidate solutions through recombination and mutation over generations. While powerful in their exploratory capability, GAs are often computationally expensive, especially when fitness evaluations are costly. These observations highlight that DKL and GA offer complementary advantages: DKL supports rapid, data-efficient exploration, while GA enables unconstrained generation of novel candidates. In this study, we combine these strengths through a DKL-GA framework that selectively evaluates a small fraction of generated candidates using active learning, thereby achieving computational efficiency without compromising the breadth of exploration.

**Method\Model**

As a computationally light model for our study, we explore the optimization of ferroelectric polarization in the processing space of a kinetic ferroelectric model FerroSIM. We stress that while sufficiently computationally light, this model provides proof-of-concept for much more complex optimization and discovery problems by providing process-trajectory dependent functional output. Originally proposed by Ricinschi et al.,(44)the FerroSIM framework provides a representation of a ferroelectric material, modeling the polarization at each lattice site as a continuous variable. This approach allows for a detailed examination of the ferroelectric domain behaviors across a discretized lattice, ensuring broad applicability to various ferroelectric systems.

The main elements of the model are as following. The local free energy at each lattice site is characterized by a Ginzburg-Landau-Devonshire (GLD) formulation adapted for the tetragonal model. In this model, the components of the polarization vector, denoted as $p_x$ and $p_y$ for the $x$ and $y$ directions respectively, are treated as order parameters. The energy landscape for each lattice site is defined by Equation 1:

$$F_{ij} = \alpha_1 \left( p_{x_{ij}}^2 + p_{y_{ij}}^2 \right) + \alpha_2 \left( p_{x_{ij}}^4 + p_{y_{ij}}^4 \right) + \alpha_3 p_{x_{ij}}^2 p_{y_{ij}}^2 - E_{loc_{x_{ij}}} p_{x_{ij}} - E_{loc_{y_{ij}}} p_{y_{ij}} \quad (1)$$

This formulation captures the quadratic and quartic contributions of the polarization components through the GLD coefficients $\alpha_1$ and $\alpha_2$, reflecting the double-well potential typical of ferroelectric materials and leading to complex history-dependent behaviors. The cross-term coefficient, $\alpha_3$, modulates the coupling between the $x$ and $y$ components of polarization at the same lattice sites, essential for representing the anisotropic interactions in tetragonal symmetry. Each lattice site is described by two indices $i$ and $j$ referring to the two-dimensional lattice structure. The local electric field at each site $E_{loc}$ integrates the effects of external field ($E_{ext}$), depolarization field ($E_{dep}$), and any local disorder $E_d(i, j)$ and is denoted by Eq. 2:

$$E_{loc} = E_{ext} + E_{dep} + E_d(i,j) \quad (2)$$

$$E_{dep} = -\alpha_{dep} < p >, \; where < p >= \textstyle\sum_N p \quad (3)$$

The depolarization field employed by the model operates through a spatially uniform depolarization factor ($\alpha_{dep}$), although this simplification is subject to the assumption that these fields are relatively constant throughout the material and is given by Equation 3. Additionally, while setting up the model, the disorder can be introduced into the system through $E_d$ by picking the sites and the intensity of the disorder to be applied. The total free energy is characterized by the sum of the free energies at individual sites and the Ising-like spin coupling between neighboring lattice sites, as expressed in Equation 4:

$$F = \sum_{i,j}^{N} F_{ij} + K \sum_{k,l} (p_{x_{ij}} - p_{x_{i+k,j+l}})^2 + K \sum_{k,l} (p_{y_{ij}} - p_{y_{i+k,j+l}})^2 \quad (4)$$

The coupling effect for the continuous order parameter, i.e., the polarization vector, is introduced as the sum of squared differences between the $x$- and $y$- components of the neighboring sites. The strength of these interactions is determined by the user-selected constant, $K$. For the scope of this study, only the nearest neighbor interactions are considered. However, incorporating interactions between next-nearest neighbors and beyond is straightforward to implement. The temporal evolution of each spin's order parameter is governed by the Landau-Khalatnikov equation, as given by Equation 5:

$$\frac{\gamma dp_{i,j}}{dt} = -\frac{\partial F}{\partial p_{i,j}} \quad (5)$$

At each time step and for each lattice site, the derivative of the total free energy ($F$) with respect to the polarization at that lattice site ($p_{i,j}$) is calculated and integrated over time to determine the evolution of the polarization vectors. Detailed explanation of the workings of the FerroSIM model can be found at these references. (38, 45)

Here, we configure a 20*20 lattice in the FerroSIM model, introducing random field disorders in either the $x$ or $y$ directions at 15% of randomly selected sites. The intensity and positions of these disorders are held constant throughout the manuscript to ensure reproducibility of the results. The input to the FerroSIM system is a time-varying electric field, applied over 3 seconds across 900 timesteps, to investigate the evolution of polarization. For simplicity, the external electric field is applied only in the x-direction and is governed by Equation 6:

$$Aexp(\alpha t) \sin(\omega t) + B \quad (6)$$

The parameters $A$, $\alpha$, and $\omega$ that govern the applied electric field are randomly sampled from uniform distributions from the intervals $A \in [0, 0.75]$, $\alpha \in [-2.75, 2.75]$, $\omega \in [-2.75, 2.75]$. These curves are normalized to fall within the range [-1, 1] and are then multiplied by a constant to ensure physical relevance to the model. Subsequently the system is equilibrated for an additional 50 timesteps at a constant electric field equal to the value at the 900th timestep.

The output we aim to optimize in this scenario is the sum of the absolute local curl at each lattice site, hereafter referred to simply as 'curl'. Curl is an unstable quantity due to the presence of applied electric field and the Ising-like spin coupling within the system. Any remnant curl decays in the equilibration phase under the influence of the constant electric field. Multiple confounding factors influence the evolution of curl in FerroSIM over time, leading to several distinct optimal solutions.(38) Thus, maximizing curl at the end of the 950 timesteps, including the equilibration period, presents an intriguing and complex challenge.

FerroSIM model offers a convenient example of the expensive physical model where the instant values of the parameters such as polarization or curl depend on the past field history representing an ideal use case for process optimization. However, for readers who are not familiar with ferroelectrics or analogous models, the FerroSIM model can be treated as representing any non-equilibrium, black-box system. In this model, the input is a high-dimensional trajectory (usually but not always a function of time) with the objective to optimize a scalar output of the system. Previously, we demonstrated the use of VAE-BO (38) and DKL (40, 41) methods for process optimization. However, these methods have limitations in deliberately generating new, optimal trajectories. In this work, we introduce a model that combines Genetic Algorithms with DKL (GA-DKL) to accelerate process optimization.

## Results

### I. Implementation of GA-DKL

In our current implementation of GA-DKL, the trajectories or the inputs to the complex physical functions serve as the chromosomes to the GA. The fitness function of each chromosome is the quantity we aim to optimize whose ground truth value is obtained by querying the high time/resource consuming physical system. As with most genetic algorithms, our goal is to maximize the fitness function, guiding the evolution of the trajectories over successive generations. This iterative process ensures that only the most promising trajectories are carried forward, continually refining the population towards optimal solutions. From here on, in the context of the current work, the words trajectories/chromosomes/electric fields all point to the input of the complex system.

In any implementation of the genetic algorithm, the evolution of chromosomes across generations is governed by operators such as crossover, mutation etc. It is important to note that the genes in our chromosomes are not independent, as our chromosomes represent electric fields that vary as a function of time or some other physical quantity. Therefore, any operators that result in non-physical (discontinuous or non-differentiable) chromosomes are avoided. To address this, we employ specialized operators that ensure the continuity and differentiability of the electric field curves. For instance, crossover operators are designed to combine parent chromosomes in a way that maintains smooth transitions between genes, avoiding abrupt changes that could render the trajectories non-physical. Mutation operators are similarly constrained to introduce variations that are gradual and preserve the overall structure of the trajectory. These operators, though less common in the literature, are crucial for maintaining the physical relevance of the solutions when applying genetic algorithms to physical systems whose inputs are trajectories. By carefully selecting and designing these genetic operators, we ensure that each generation of chromosomes evolves towards more optimal solutions without violating the physical constraints of the system. It is however important to note that other complex operators that might result in non-physical systems can also be incorporated without modifying the GA-DKL method discussed in this manuscript.

In our implementation of the genetic algorithm, three key operations govern the evolution of chromosomes across generations: carry-over, crossover, and mutation. First, the carry-over operation ensures that a pre-determined number of chromosomes with the highest fitness values are preserved and directly carried over to the next generation. This elitist strategy guarantees that the best solutions are retained, providing a strong foundation for further optimization. For the crossover operation, we employ a variation known as arithmetic crossover. Given two parent chromosomes, $par_1$ and $par_2$, we form two child chromosomes, $chd_1$ and $chd_2$, using Equation 7:

$$
\begin{aligned}
chd_1 &= \lambda^* * par_1 + (1-\lambda^*) * par_2 \\
chd_2 &= (1-\lambda^*) * par_1 + \lambda^* * par_2
\end{aligned} \tag{7}
$$

This method involves a linear combination of the two parent curves, where $\lambda^*$ is randomly sampled from the range [0, 1], determining the influence of each parent chromosome on its offspring. This approach ensures that the children are both continuous and differentiable, adhering to the physical constraints of the system. A few examples of this operation are shown in Figure 1a, where the parents are chosen randomly from the FerroSIM case and are depicted in blue, while the two child chromosomes formed for each set of parents are shown in red. The value of $\lambda^*$ for each case is indicated on top of the individual plots. When $\lambda^*$ is too small or too large, the children closely resemble one of the parents. In other cases, the child chromosomes inherit characteristics from both parents.

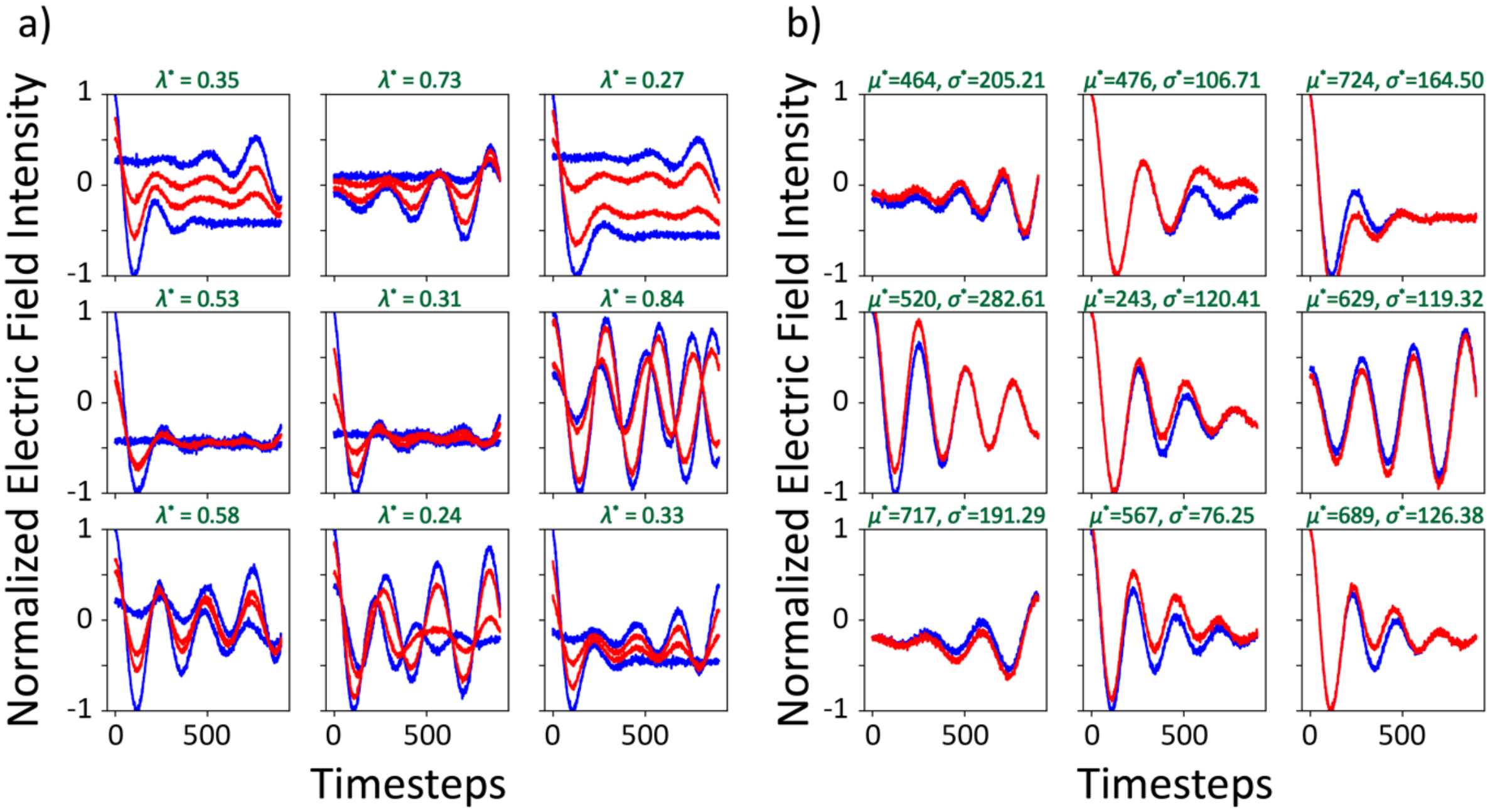

**Figure 1.** Examples of genetic operators used in the GA for continuous input spaces. Parent chromosomes are shown in blue which are randomly selected from the FerroSIM input dataset.

Children chromosomes are shown in red and are generated by arithmetic crossover in (a) and mutation in (b). The weights ($\lambda^*$) that formed the children chromosomes in arithmetic crossover (a) and the parameters ($\mu^*$ and $\sigma^*$) that formed the children in case of mutation (b) are shown on top of each subplot.

The mutation operation modifies the parent chromosome by adding a weighted Gaussian function and is represented by Equation 8. In the context of the GA operators discussed here, the chromosomes are normalized to fall in the range [-1, 1] as seen by GA and when needed to evaluated for a fitness function are multiplied by a constant to make them relevant for the FerroSIM system. Each chromosome is divided into 900 timesteps and each timestep constitutes a gene. Specifically, for each parent chromosome, we select a mean ($\mu^*$) from a uniform distribution within the range [100, 800] and a standard deviation ($\sigma^*$) from a half-normal distribution with a location of 50 and a scale of 150. We then generate a Gaussian probability density function (pdf) over the 900 timesteps, using the selected parameters $\mu^*$ and $\sigma^*$. A weight factor is determined by sampling a random value from a uniform distribution over [50, 150], and the Gaussian function is scaled by this weight factor ($w$) and a randomly chosen sign (+ or -). The resulting Gaussian function is then added to the parent chromosome. This operation ensures that the resulting child chromosome remains continuous and differentiable, adhering to the physical constraints of the problem. This operation is performed on randomly selected parents (shown in blue) and their corresponding children are shown in red in Figure 1b. The constants $\mu^*$ and $\sigma^*$ for each of the cases are shown on top of the subplots in Figure 1b.

$$chd = par \pm \left(w * pdf(\mu, \sigma)\right) \tag{8}$$

The key step that we introduce in this work, in the context of accelerating genetic algorithms is the usage of the DKL model for quick fitness function estimation which determines the selection of candidates for the next iteration. In classical GA methods, the selection process determines which individuals from the current population will contribute to the next generation. The greediest selection method is elitism, where the best-performing individuals are directly carried over to the next generation, ensuring that the highest quality solutions are always preserved. Moving towards more exploratory methods, roulette wheel selection assigns selection probabilities proportional to individuals' fitness, allowing even less fit individuals a chance to be selected based on their relative performance. Tournament selection introduces a competitive element by randomly selecting a subset of individuals and choosing the best among them, balancing exploration and exploitation. Finally, rank-based selection sorts individuals by fitness and assigns selection probabilities based on their ranks rather than their absolute fitness values, promoting diversity and reducing the dominance of highly fit individuals. These varying degrees of greediness and exploration in the selection process help maintain a balance between refining existing solutions and exploring new areas of the search space. However, all these methods require the evaluation of the fitness function for tens of thousands of chromosomes across generations, which, when computationally expensive, can severely limit the applicability of genetic algorithms in cases like process optimization. In this study, we address the challenge of accelerating the evaluation of the fitness function in each generation, thereby enhancing the efficiency of the genetic algorithm.

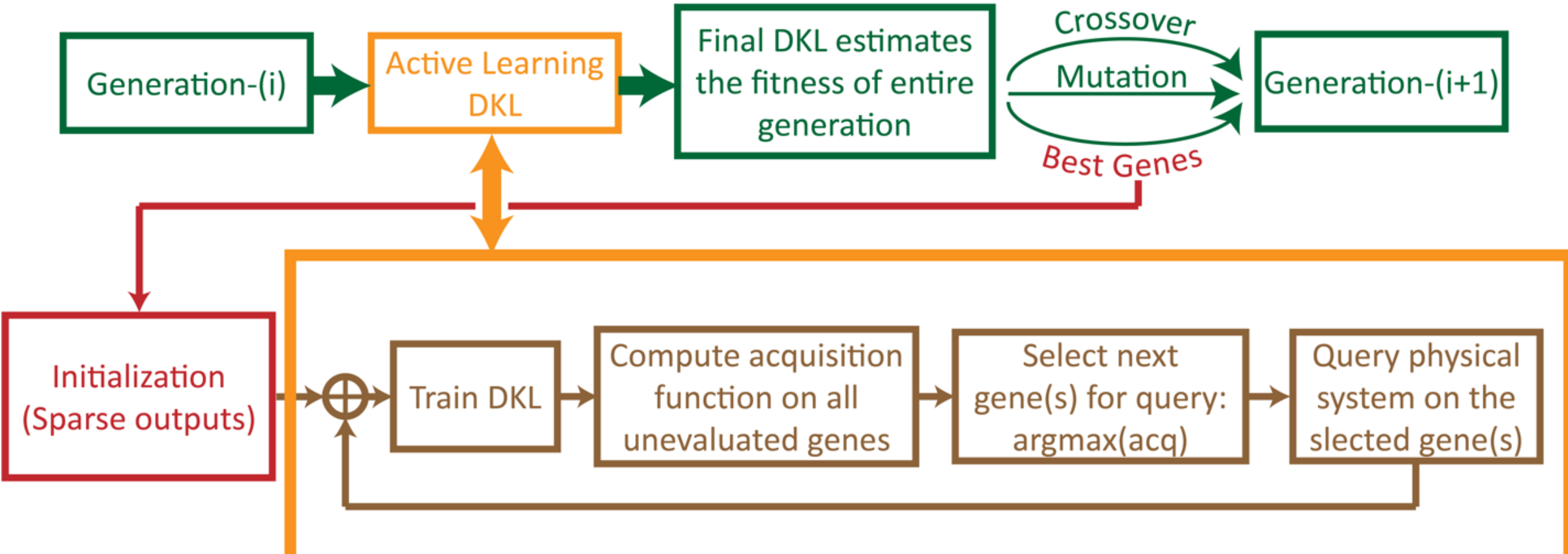

**Figure 2.** Workflow of the DKL-GA method, where the top row shows the how the next generation of chromosomes are created using fitness function values of the existing generation. The process of estimating fitness function values is accelerated by running an active learning loop of DKL in the current generation and is shown in the bottom row in orange.

To accelerate the GA algorithm, we use deep kernel learning (DKL) to actively search the space of possible trajectories within each generation constructed by the genetic algorithm. DKL is a method that combines the power of deep neural networks with the flexibility of Gaussian Processes (GPs) to model and make predictions on complex target functions in high-dimensional spaces. The ability to efficiently explore the high-dimensional input space and learn the relationship between the inputs and target function, while quantifying the uncertainties in the predictions (due to the presence of GP), makes it a powerful tool for optimization and prediction tasks. We note that DKL is an active learning method associated with its own policies that can vary from purely exploratory to very greedy. We discuss the effects of combining different GA and DKL selection policies in forthcoming sections.

At each generation, a small number of chromosomes are randomly selected (e.g., 1%), and their corresponding ground truth fitness values are evaluated by querying the physical system. Further acceleration of the initialization step is achieved by constructing the training dataset using the evaluated chromosomes that got carried over from the previous generation. This set of chromosomes and fitness function values is used as the initial training dataset for a DKL network. The trained DKL network is then used to estimate the fitness function values of the remaining chromosomes in the model. This step can be implemented either as a one-time estimation, or DKL can be used to actively explore the candidate space in the usual active learning form. In both cases, DKL acts as a surrogate model that allows predicting target functionality and its uncertainty for a full set of chromosomes, substituting the expensive evaluation of the full generation.

For the former case (static learning), the DKL's training dataset is picked randomly and might not result in a 'good-enough' fitness function estimation. In the latter case (active learning), DKL intelligently selects chromosome(s) based on its predictions and uncertainty estimations via a chosen acquisition function. These evaluated chromosome(s) and their corresponding fitness function values are then added to the initial dataset, completing one iteration of the active-learning scenario using DKL. This active learning is run for a user selected number of iterations or until a stop condition is met and the final DKL model that is trained on the augmented training dataset is then used to estimate fitness function values on the remaining population. The entire workflow of implementing DKL to actively learn a generation in GA and use it to estimate the values of the remaining population without actually querying the time-consuming physical system for the entire population is shown in Figure 2.

The acquisition function balances the exploration-exploitation trade-off of the active learning, allowing the algorithm to cautiously explore the search space (generation) while also prioritizing the optima. The active learning scenarios using DKL are thoroughly discussed in references where detailed explanations of how DKL outperforms other similar algorithms for high-dimensional process optimization tasks are also available.(40, 41) Each generation is actively learned by the DKL, with chromosomes to be queried diligently selected until only a portion (e.g., <10%) of the generation is explored. It is important to note that the DKL trained on this dataset at the end of the active learning has a thorough understanding of the optima and the search space within generation, whereas GA drastically changes the population via mutation and cross-over from generation to generation.

## II. GA-DKL on FerroSIM

To illustrate the GA-DKL approach, we apply it to the FerroSIM system, which serves as an example of a physical system where evaluating the fitness function for all genes is impractical. In our implementation of the GA, we start with 1000 electric fields in generation-0, each divided into 900 timesteps that act as chromosomes for the GA implementation. This initial set is a subset of the training dataset used in previous studies, ensuring that it represents a diverse range of potential solutions.(38, 40) The electric field is (a) applied only in the x-direction, (b) includes an equilibration region of 50 timesteps of constant electric field at the end, and (c) normalized to fall in the range [-1, 1] as seen by GA and then multiplied by a constant to make it relevant to the physical system. The ground truth fitness function of each trajectory is obtained by running a FerroSIM simulation with the given electric field (and equilibration) and calculating the property of interest, specifically the curl, at the end of the simulation. In this context, the fitness function reflects the ability of each electric field curve to achieve the desired physical outcome. By quantifying the performance of each trajectory, we can effectively compare and rank them. As discussed in the previous section, we implement three operations to evolve our chromosomes with the aim of improving the fitness function across generations. A predetermined number (15%) of chromosomes, sorted by fitness function values, are carried over to the next generation directly. Either crossover or mutation operations are then applied to create offspring chromosomes until about 1000 chromosomes are generated for the next generation.

At the start of each generation, the physical system is queried for a small number of chromosomes to obtain the ground truth fitness function values for these chromosomes. These chromosome-ground truth fitness function pairs are used as the initial dataset for the DKL network. In our implementation, we select the chromosomes that are carried over from the previous generation and evaluate them for the ground truth fitness as the initial dataset to avoid initial querying. Once the DKL is trained on the initial dataset, the next chromosome(s) to be queried is selected based on the upper confidence bound (UCB) acquisition function of the form $\mu + \xi * \sigma$, where $\xi$ provides direct control over the exploration-exploitation trade-off involved with Bayesian Optimization (BO). The value of $\xi$ is set to 10 to ensure that the BO explores the entire generation to query the chromosomes expected to have higher fitness function values ($\mu$) or that may have high fitness value due to high uncertainty ($\sigma$).

In the implementation of this algorithm for the FerroSIM scenario, at every iteration of BO, we select a batch of 10 chromosomes with the highest values of the acquisition function. The parallel simulations are run for the selected batch on a 10-core CPU at a time to further accelerate the process. The BO is run until it queries 100 chromosomes in each generation, equating to merely 10% of the chromosomes queried in each generation. This acceleration is compounded by the fact that 10 queries at every iteration of DKL take similar time to one query run on a single core CPU. The dataset used to train the DKL at the end of the exploration comprises the initial chromosomes carried over from the previous generation and the chromosomes selected by DKL at every iteration that have high values of the acquisition function. This dataset provides a comprehensive overview of the entire generation to the DKL since the selected chromosomes either minimize the uncertainty in the predictions or are expected to have high fitness function values. At the end of each exploration, the fitness function of all the chromosomes in the generation are estimated by the DKL at the end of the exploration.

At the end of the exploration phase, we estimate the fitness function values for all chromosomes in the generation by sampling from the DKL's posterior function, similar to Thompson sampling in BO. Note that this policy selection to estimate fitness function from DKL's predictions is somewhat arbitrary and will be further discussed later in the manuscript. This sampling is performed before each crossover and mutation operation. The likelihood of selecting a chromosome as a parent for these operations is proportional to its fitness function value, as sampled from the DKL's posterior. This approach ensures that the DKL's predictions and associated uncertainties are considered for estimating the fitness function values, thereby guiding the selection process more effectively.

The results of implementing the GA-DKL on the FerroSIM system are shown in Figure 3, where in plots (a-e), the distribution of the target property in generation-0 is shown in red. The GA algorithm is run for 40 iterations and the distributions of ground truth fitness functions values of queried chromosomes after every $10^{th}$ generation are shown in blue in Figures. 3a-e. These queried chromosomes by the DKL as discussed provide a comprehensive overview of the generation. The maximum value of the fitness function at each generation among all the explored chromosomes are plotted in Figure. 3e. As shown from previous studies (40, 41, 46), the DKL is expected to explore the optima of the dataset well before exploring 10% of the dataset for complex problems where other optimization problems fail.

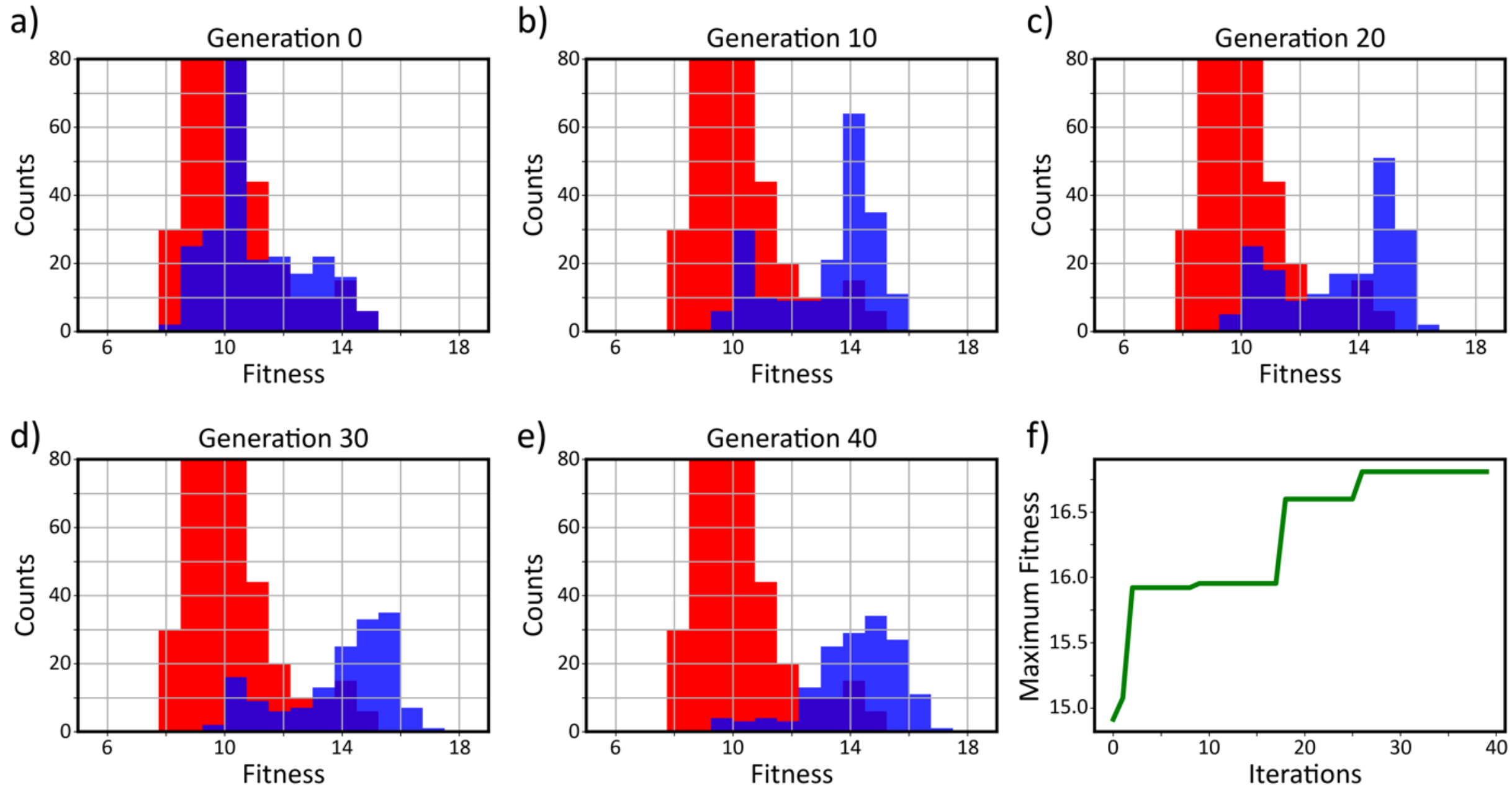

**Figure 3.** Results of 40 iterations of GA-DKL on FerroSIM system. Ground truth distribution of curl (fitness function) of all the chromosomes in generation-0 are shown in red (a-e). Ground truth curl distribution of explored points by DKL in generation (a) 0, (b) 10, (c) 20, (d) 30, and (e) 40 is shown in blue. The maximum value of the fitness values among the chromosomes explored by DKL as a function of generation is shown in (e).

The results of the study demonstrate that as the genetic algorithm (GA) progresses through generations, the distributions of the fitness functions shift to the right. This shift occurs because weaker individuals are systematically eliminated, while stronger genes are favored and carried forward. Consequently, this process leads to the generation of better input curves, or trajectories, that were not present in the initial dataset. This phenomenon is visually illustrated in Figure 3e, where the carryover operation preserves the best genes in the current population, while the other operations introduce better genes at regular intervals, which is a characteristic feature of genetic algorithms.

## III. Policy effect on the GA-DKL

In the current implementation of our GA-DKL algorithm, two policies play a pivotal role in determining its overall efficacy. The first policy involves selecting the acquisition function for the active learning phase of DKL, which governs the training dataset used to estimate the fitness of unqueried chromosomes. This training dataset is crucial as it determines the accuracy of DKL's predictions for the rest of the population, thereby influencing the chromosomes' chances of participating in GA operations. Ideally, the training dataset for DKL should balance the exploration-exploitation trade-off—a typical property of optimization algorithms—minimizing uncertainty while thoroughly exploring the optima within the generation. If DKL's exploration becomes confined to a local or global optimum, it could lead to inaccurate predictions for the remaining population, adversely affecting the next generation. Conversely, an overly exploratory active learning process could misrepresent the optimal chromosomes, limiting their potential contributions to the next generation.

To evaluate different acquisition functions within the GA-DKL context, we selected the 10th generation from the 40 explored generations in our study as a qualitative litmus test for assessing DKL's exploration efficacy in active learning scenarios. Initially, we queried the physical system for all chromosomes in this generation, obtaining the ground truth fitness function values. This step allows us to measure the effectiveness of each acquisition function in exploring the generation. Over 10 iterations, DKL explored this generation by querying the physical system for the ten best chromosomes per iteration, as determined by the maximum acquisition function values. We employed five different acquisition functions for this analysis: (a) mean or greedy exploration, (b) uncertainty-based exploration, (c) upper confidence bound (UCB), (d) expected improvement (EI), and (e) probability of improvement (POI). The outcomes of these explorations are illustrated in Figure 4, where the model's predicted versus actual fitness function values are plotted at the end of the exploration; the points DKL selected for querying are highlighted in red for each case. Additionally, the root mean square error (RMSE) of the predictions across the entire generation and the average uncertainty of DKL's predictions over the generation are displayed in each plot. An ideal acquisition function should evenly explore the search space while also effectively sampling the optima. This would result in lower RMSE over the entire generation while maintaining high prediction accuracy near the optima. The ability to accurately capture the optima is critical in the GA-DKL context, as these are the chromosomes most likely to be selected as parents for the next generation. Any significant errors in the prediction of these high-performing candidates may lead to the construction of sub-optimal offspring, ultimately hindering the convergence of the genetic algorithm toward global optima.

Greedy exploration, illustrated in Figure 4a, considers only the mean of the DKL's predictions to select the next chromosome for querying, focusing all resources on exploring the optima at every iteration. However, this approach leads to insufficient exploration of the rest of the search space, resulting in a high root mean square error (RMSE) across the entire generation. Additionally, there is a significant risk of the exploration becoming confined to a local optimum. In contrast, uncertainty-based exploration, as evident in Figure 4b, aims to cover the entire space, often without specifically targeting the optima, which results in only a few chromosomes near the optima being explored. The UCB-based exploration, shown in Figure 4c, strikes a balance between exploration and exploitation by targeting the optima while still covering the entire generation. This trade-off is controlled by the hyperparameter $\lambda$, set at 10, as discussed in the previous section. Ideally, the training dataset for DKL's exploration of the GA generation should contain information about both the optima and the broader dataset to ensure a comprehensive exploration of the search space. This is reflected in the exploration results of UCB, where more points near the maximum are sampled than in uncertainty-based exploration and a lower RMSE for the entire generation is achieved compared to mean-based exploration. Additionally, the results of the exploration using EI and POI are shown in Figures 4d and 4e, respectively. These acquisition functions, along with UCB, are known to balance the exploration-exploitation trade-off, resulting in a robust training dataset for DKL to estimate the fitness function of the unqueried chromosomes. The parameters controlling the trade-off in EI and POI have not been optimized for this analysis and are only discussed here as potential alternatives to the acquisition function used in this study (UCB). The exploration with optimized parameters is expected to follow a similar trend to that of UCB for both cases.

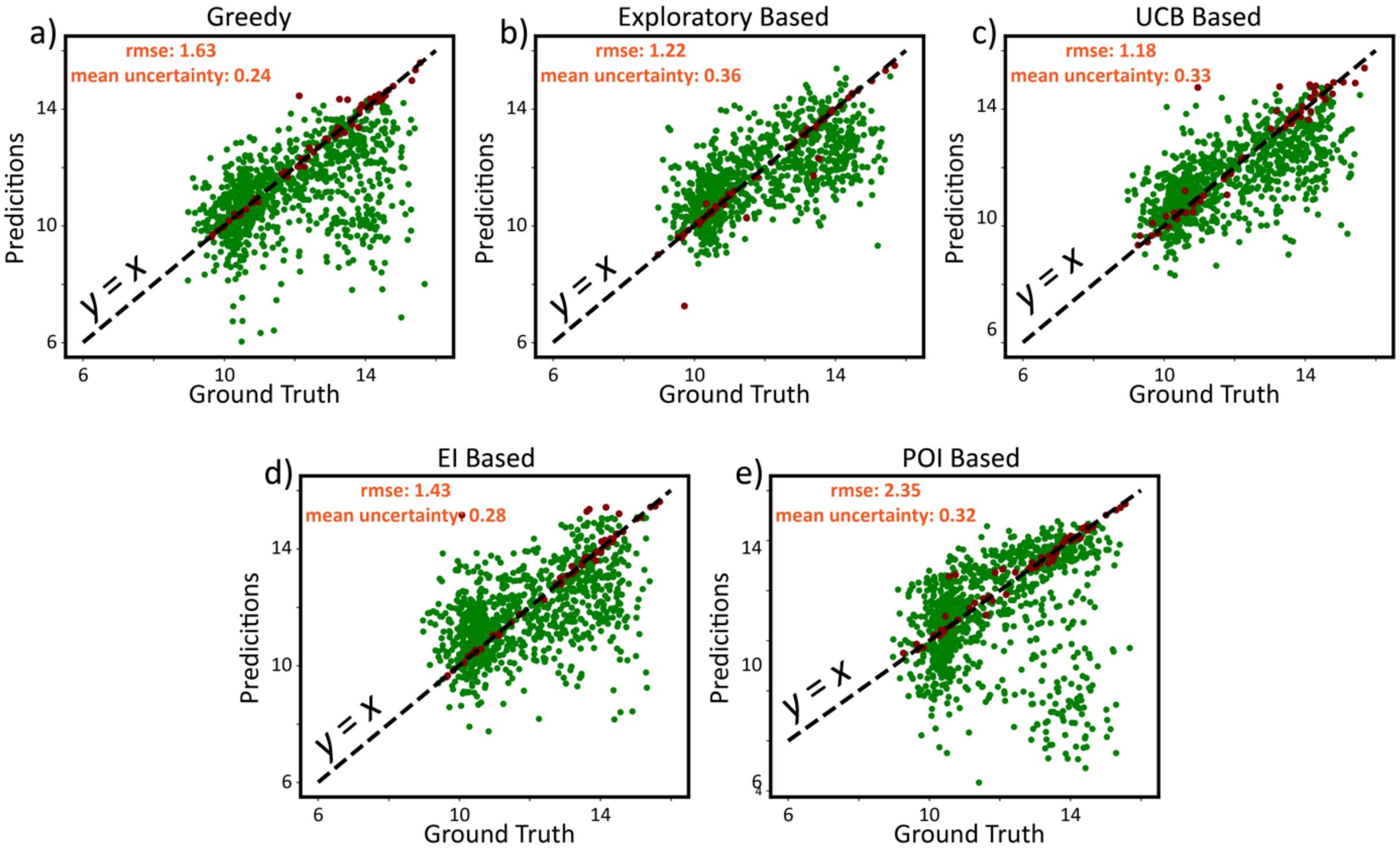

**Figure 4.** Effect of different acquisition functions on active learning within the DKL-GA framework, demonstrated using an arbitrary (10th) generation of chromosomes. Each subplot shows the predicted versus ground truth fitness values after 100 queries guided by a specific acquisition function: (a) mean-based, (b) uncertainty-based, (c) UCB, (d) EI, and (e) POI. Red points indicate chromosomes selected for querying; green points represent those whose fitness values were predicted. The RMSE and the average uncertainty over the entire generation at the end of the exploration is shown in each plot. An ideal acquisition function should explore not only the optima but also the broader landscape of the fitness function—sampling from maxima, minima, and intermediate regions. This ensures that the resulting predictions on the unexplored chromosomes are as close as possible to the ground truth, with minimal uncertainty.

The second policy explored here governs the estimation of fitness function values for unqueried chromosomes in the generation, using actively learned DKL. The predictions from DKL are obtained in the form of normal distributions, governed by parameters such as the mean and the covariance matrix. This covariance matrix allows us to estimate the uncertainty of the predictions, a characteristic inherent to Bayesian methods. In our implementation discussed in the previous section, at the end of DKL's exploration phase, the fitness function values are sampled using Thompson Sampling, where both mean, and uncertainty are utilized in the estimation. This process is conducted before selecting the parents for each operation while forming the next generation. However, in this section, we study three different methods of estimating fitness function values using the actively trained DKL: mean-only, uncertainty-only, and Thompson Sampling. The acquisition function UCB is used as policy-1 all three cases to maintain consistency for this exploration. Further to maximize the discrepancies between the policies, the capacity of the algorithm is decreased wherever possible. For example, the number of convolutional filters in the neural network of the DKL are decreased from 128 (in previous sections) to 32. Along with that, at each iteration of active learning, only one chromosome is sampled and sent for querying as opposed to sampling 10 for parallel querying. This restricts the number of chromosomes queried by the algorithm in each generation to merely 10 chromosomes (1 of each generation).

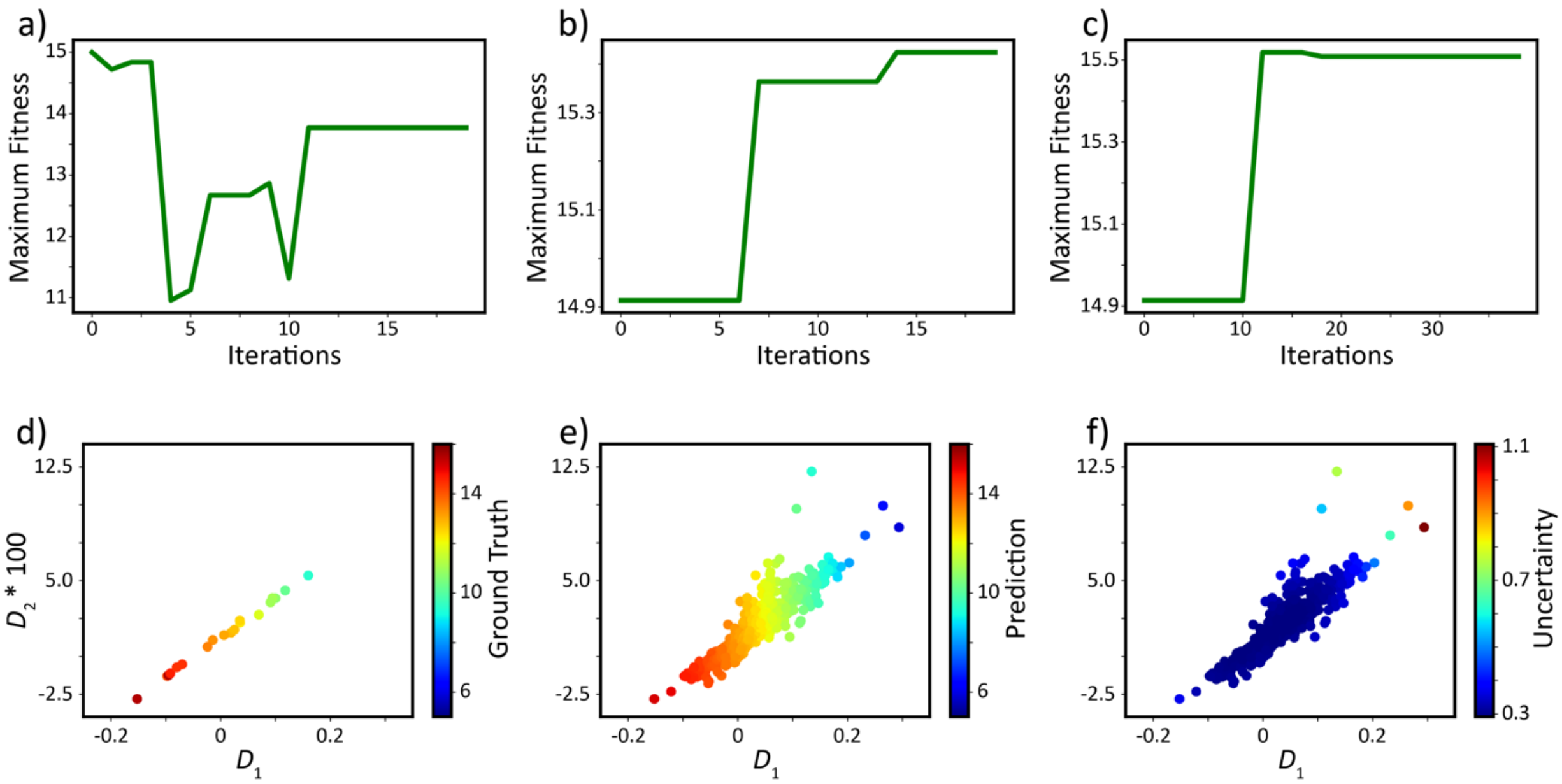

**Figure 5.** Comparison of three different strategies for estimating fitness values of unqueried chromosomes in the DKL-GA framework: (a) uncertainty-only, (b) mean-only, and (c) Thompson sampling. Each plot shows the highest ground truth curl value among the queried chromosomes as a function of the generation, illustrating how well each estimation method guides selection of high-performing chromosomes over time. For a more detailed view, subfigures (d)–(f) show a 2D latent representation of all chromosomes from a representative (10th) generation with Thompson sampling employed as the second policy, colored by (d) ground truth curl values of the explored genes, (e) DKL-predicted values of all genes, and (f) predictive uncertainty of all genes. The output layer of the neural network in the trained DKL at the end of exploration in each generation, forms these low-dimensional representation of chromosomes.

Uncertainty-based exploration proves the least effective among the three cases, as the UCB acquisition function more adeptly explores the optima, resulting in lower uncertainties associated with chromosomes near these points. The highest ground truth curl values among the chromosomes explored during the active learning phase for this case are shown in Figure 5a. Mean-based exploration (Figure 5b) which disregards the uncertainty in the predictions, appears to generate optimal curves at an equivalent pace to that of Thompson sampling, as demonstrated in Figure 5c. We note the sudden jump in case of Thompson sampling, nonetheless, the maximum value of the fitness function at the end of the exploration are comparable in the Figure. 5b and 5c. These results indicate that, within the context of this system, incorporating uncertainty into the prediction of fitness function values does not lead to the construction of better generations. This is because the elevated uncertainties do not appear to correspond with the optimal chromosomes, thereby reducing their relevance for parent selection. This is evidenced by the low-dimensional representation of chromosomes at the end of the randomly selected generation ($10^{th}$) of the Thompson policy, colored according to ground truth (Figure 5d), predictions of the DKL (Figure 5e), and uncertainty associated with predictions (Figure 5f). The embedding space of the trained DKL forms the low dimensional space shown in Figures. 5d-f. The ground truth and predictions indicate that DKL's predictions are accurate even with 1% exploration, and the uncertainties are elevated in the unexplored regions of the latent space, which in our case are not associated with the optima. This appears to be the general trend for all generations, thereby diminishing the role of uncertainty in estimating the fitness function. However, Thompson sampling is expected to perform better when the information from uncertainty estimation becomes crucial. For instance, when multiple optima are not thoroughly explored, this can result in high uncertainty for chromosomes near these optima. Additionally, if some of the curves diverge from the general trend of the dataset, they will exhibit higher uncertainties and may play a crucial role in forming better chromosomes for the next generation. These scenarios are beyond the scope of the current system: FerroSIM used in this study. However, it is advisable to use both means and uncertainties in selecting the fitness function for more complex datasets, as this approach is generally safer.

**Summary**

We propose the GA-DKL as an innovative approach to exploring complex multidimensional optimization spaces. This method leverages the universality and robustness of Genetic Algorithms (GA) combined with the efficiency of Deep Kernel Learning (DKL) to navigate and optimize intricate solution landscapes effectively. Compared to the DKL and VAE-BO based methods this approach brings the generative element effective for out-of-distribution exploration, whereas compared to pure DKL it allows considerably higher efficiency since candidate solutions can now be evaluated only for a fraction of the candidate space.

Here, this approach was applied to the FerroSIM system to investigate the impact of different policies and seed values on the optimization process. The GA-DKL method was able to efficiently explore and optimize the system, demonstrating its applicability to various scenarios. In particular, the algorithm has clearly proposed the solutions that have not been a part of the original candidate space and offer improvements in target function.

We note that the proposed DKL-GA solution is ideally suited for the exploration of both differentiable and non-differentiable spaces, such as molecular discovery and other complex scientific problems.

**Acknowledgements**

The GA-DKL concept and workflow development (M.V., A.R., S.V.K) is supported by the U.S. Department of Energy, Office of Science, Office of Basic Energy Sciences as part of the Energy Frontier Research Centers program: CSSAS-The Center for the Science of Synthesis Across Scales, under award number DE-SC0019288.